\documentclass{article}
\usepackage{amsmath,amsfonts,bm}

\def\eqref#1{equation~\ref{#1}}
\def\1{\bm{1}}

\def\vx{{\bm{x}}}

\DeclareMathAlphabet{\mathsfit}{\encodingdefault}{\sfdefault}{m}{sl}
\SetMathAlphabet{\mathsfit}{bold}{\encodingdefault}{\sfdefault}{bx}{n}

\newcommand{\E}{\mathbb{E}}

\PassOptionsToPackage{numbers, compress}{natbib}
\usepackage[preprint]{neurips_2022}

\usepackage[utf8]{inputenc} % allow utf-8 input
\usepackage[T1]{fontenc}    % use 8-bit T1 fonts
\usepackage{hyperref}       % hyperlinks
\usepackage{url}            % simple URL typesetting
\usepackage{booktabs}       % professional-quality tables
\usepackage{amsfonts}       % blackboard math symbols
\usepackage{nicefrac}       % compact symbols for 1/2, etc.
\usepackage{microtype}      % microtypography
\usepackage{xcolor}         % colors

\definecolor{rubinered}{RGB}{234,52,35}
\definecolor{olivegreen}{RGB}{55,126,33}

\newtheorem{theorem}{Theorem}
\newtheorem{lemma}{Lemma}

\usepackage{mathtools}
\usepackage{amssymb}
\usepackage{multirow}
\usepackage{subcaption}
\usepackage{wrapfig}
\usepackage{algorithm}
\usepackage{algpseudocode}

\title{Hazard Gradient Penalty for Survival Analysis}

\author{%
  Seungjae Jung \\
  NAVER R\&D Center\\
  NAVER CLOVA\\
  \texttt{seung.jae.jung@navercorp.com} \\
   \And
   Kyung-Min Kim \\
   NAVER CLOVA \\
   NAVER AI LAB \\
   \texttt{kyungmin.kim.ml@navercorp.com}
}

\begin{document}

\maketitle

\begin{abstract}
Survival analysis appears in various fields such as medicine, economics, engineering, and business.
Recent studies showed that the Ordinary Differential Equation (ODE) modeling framework unifies many existing survival models while the framework is flexible and widely applicable.
However, naively applying the ODE framework to survival analysis problems may model fiercely changing density function which may worsen the model’s performance.
Though we can apply L1 or L2 regularizers to the ODE model, their effect on the ODE modeling framework is barely known.
In this paper, we propose \textit{hazard gradient penalty} (HGP) to enhance the performance of a survival analysis model.
Our method imposes constraints on local data points by regularizing the gradient of hazard function with respect to the data point.
Our method applies to any survival analysis model including the ODE modeling framework and is easy to implement.
We theoretically show that our method is related to minimizing the KL divergence between the density function at a data point and that of the neighborhood points. 
Experimental results on three public benchmarks show that our approach outperforms other regularization methods. 
\end{abstract}

\section{Introduction}

Survival analysis (a.k.a time-to-event modeling) is a branch of statistics that predicts the duration of time until an event occurs \cite{kleinbaum2012survival}.
Survival analysis appears in various fields such as medicine \cite{schwab2021real}, economics \cite{meyer1988unemployment}, engineering \cite{oconner2011reliability}, and business \cite{jing2017neuralsurvival, li2021churn}.
Due to the presence of right-censored data, which is data whose event has not occurred yet, survival analysis models require special considerations.
Cox proportional hazard model (CoxPH) \cite{cox1972regression, katzman2018deepsurv} and accelerated time failure model (AFT) \cite{wei1992accelerated} are widely used to handle right-censored data.
Yet the assumptions made by these models are frequently violated in the real world.
Recent studies showed that the Ordinary Differential Equation (ODE) modeling framework unifies many existing survival analysis models including CoxPH and AFT \cite{groha2020survnode, tang2022survival}.
They also showed that the ODE modeling framework is flexible and widely applicable.

However, naively applying the ODE framework to survival analysis problems may result in wildly oscillating density function that may worsen the model’s performance.
Regularization techniques that can regularize this undesirable behavior are understudied.
Though applying L1 or L2 regularizers to the ODE model is one option, their effects on the ODE modeling framework are barely known.
The cluster assumption from semi-supervised learning states that the decision boundaries should not cross high-density regions \cite{chapelle2006semi}.
Likewise, survival analysis models need hazard functions that slowly change in high-density regions.

In this paper, we propose hazard gradient penalty to enhance the performance of a survival analysis model.
In a nutshell, the hazard gradient penalty regularizes the gradient of the hazard function with respect to the the data point from the real data distribution.
Our method has several advantages.
1) The method is computationally efficient.
2) The method is theoretically sound.
3) The method is applicable to any survival analysis model including the ODE modeling framework as long as it models hazard function.
4) It is easy to implement.
We theoretically show that our method is related to minimizing the KL divergence between the density function at a data point and that of the neighborhood points of the data point.

Experimental results on three public benchmarks show that our approach outperforms other regularization methods.

\section{Preliminaries}

Survival analysis data comprises of an observed covariate $\vx$, a failure event time $t$, and an event indicator $e$.
If an event is observed, $t$ corresponds to the duration time from the beginning of the follow-up of an individual until the event occurs.
In this case, the event indicator $e = 1$.
If an event is unobserved, $t$ corresponds to the duration time from the beginning of follow-up of an individual until the last follow-up.
In this case, we cannot know the exact time of the event occur and event indicator $e = 0$.
An individual is said to be \textit{right-censored} if $e = 0$.
The presence of \textit{right-censored} data differentiates survival analysis from regression problems.
In this paper, we only focus on the single-risk problem where event $e$ is a binary-valued variable.

Given a set of triplet $\mathcal{D} = \{ (\vx_i, t_i, e_i) \}_{i=1}^N$, the goal of survival analysis is to predict the likelihood of an event occur $p(t \mid \vx)$ or the survival probability $S(t \mid \vx)$.
The likelihood and the survival probability have the following relationship:
\begin{equation} \label{eq:surv_likelihood_relation}
S(t \mid \vx) = 1 - \int_0^t p(\tau \mid \vx) d\tau
\end{equation}

Modeling $p(t \mid \vx)$ or $S(t \mid \vx)$ directly is challenging as those have the following constraints:
\begin{align*}
p(t \mid \bm{x}) > 0, \quad &\int_0^\infty p(\tau \mid \bm{x}) d\tau = 1 \\
S(0 \mid \vx) = 1, \quad \lim_{t \rightarrow \infty} S(t \mid \vx) =  0, \quad
& S(t_1 \mid \vx) \geq S(t_2 \mid \vx) \text{ if } t_1 \leq t_2 
\end{align*}

Previous works instead modeled the hazard function (a.k.a conditional failure rate) $h(t \mid \vx)$ \cite{cox1972regression, katzman2018deepsurv, wei1992accelerated, zhong2022deepeh}.
\begin{equation} \label{eq:hazard_func_def}
h(t \mid \vx)
\coloneqq \lim_{\Delta t \rightarrow 0} \frac{P(t \leq T < t + \Delta t\mid T \ge t, \vx)}{\Delta t}
= \frac{p(t \mid \vx)}{S(t \mid \vx)}
\end{equation}

As the hazard function is a probability per unit time, it is unbounded upwards.
Hence, the only constraint of the hazard function is that the function is non-negative: $h(t \mid \vx) \geq 0$

\subsection{The ODE Modeling Framework}

We can obtain an ODE which explains the relationship between the hazard function and the survival function by putting derivative of \eqref{eq:surv_likelihood_relation} into \eqref{eq:hazard_func_def} \citep{kleinbaum2012survival}.

\begin{equation} \label{eq:hazard_surv_ode}
h(t \mid \vx) = \frac{p(t \mid \vx)}{S(t \mid \vx)} = \frac{1}{S(t \mid \vx)} \left( - \frac{d S(t \mid \vx)}{dt}\right) = -\frac{d \log S(t \mid \vx)}{dt}
\end{equation}

Starting from initial value $\log S(0 \mid \vx) = 0$, we can define $\log S(t \mid \vx)$ as the solution of the ODE initial value problem where the ODE is defined as \eqref{eq:hazard_surv_ode}.
\begin{equation} \nonumber
\log S(t \mid \vx) = \log S(0 \mid \vx) + \int_0^t -h(\tau \mid \vx) d\tau = \int_0^t -h(\tau \mid \vx) d\tau
\end{equation}

We can train the ODE model by minimizing the negative log-likelihood.
\begin{align}
\mathcal{L}_\vx
&= - e \log p_\theta(t \mid \vx) + (1 - e) \log S_\theta(t \mid \vx) \label{eq:negative_log_likelihood} \\
&= - e \left( \log h_\theta(t \mid \vx) + \log S_\theta(t \mid \vx) \right) + (1 - e) \log S_\theta(t \mid \vx) \nonumber \\
\end{align}
Following \citet{groha2020survnode}, we update the model parameter using Neural ODE \cite{chen2018node}.
The hazard function $h_\theta (t \mid \vx)$ is modeled using a neural network followed by the softplus activation function to ensure that the output is always non-negative.

\section{Methods}

\begin{wrapfigure}{R}{0.45\textwidth}
    \centering
    \vspace{-\intextsep}
    \includegraphics[width=0.9\linewidth]{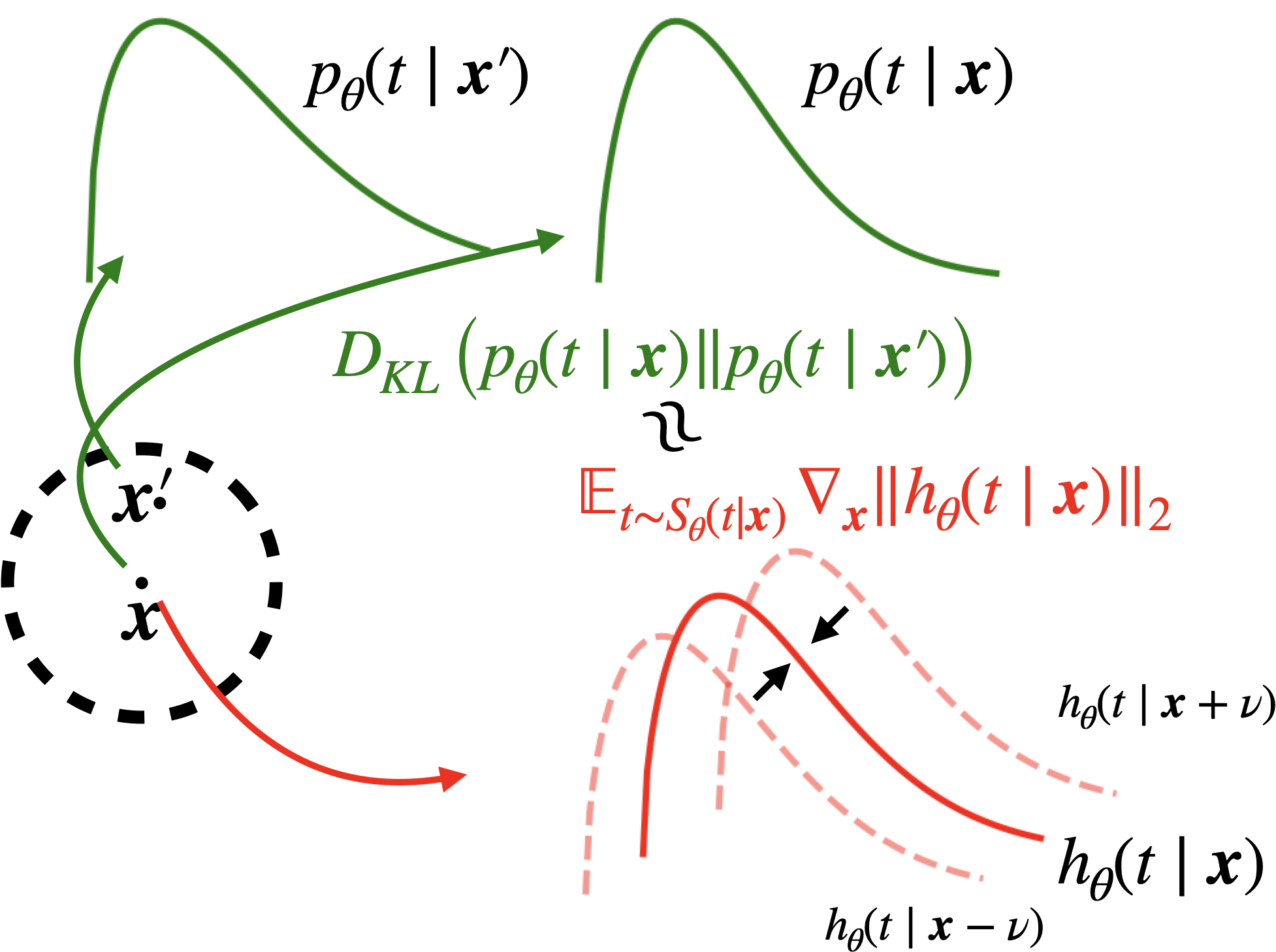}
    \caption{
        Graphical overview of our proposed method.
        Our method minimize 
        \textcolor{rubinered}{
            the hazard gradient penalty $\E_{S_\theta(t \mid \vx)} \left\Vert \nabla_\vx h_\theta(t \mid \vx) \right\Vert_2$
        }
        and the negative log-likelihood in \eqref{eq:negative_log_likelihood} at the same time.
        Intuitively speaking, we regularize the model so that the hazard function does not vary much when a small noise $\nu$ is added or subtracted to the data point $\vx$.
        In section \ref{sec:kl_div_connection}, we show that minimizing the hazard gradient penalty is connected to minimizing \textcolor{olivegreen}{the KL divergence between the density at $\vx$ and the density at $\vx^\prime \in B(\vx, \epsilon)$}.
    }
    \label{fig:hazard_gradient_penalty}
\end{wrapfigure}

In this section, we introduce the hazard gradient penalty and show that it is related to minimizing the KL divergence between the density function at a data point and that of its neighbours.
See Figure \ref{fig:hazard_gradient_penalty} for the graphical overview of our method.

The cluster assumption from semi-supervised learning states that the decision boundaries should not cross high-density regions \cite{chapelle2006semi}.
In a similar vein, hazard functions of survival analysis models should change slowly  in high-density regions.
To achieve this, we propose the following regularizer.\footnote{
In practice, we implement $h_\theta (t \mid \vx)$ using a neural network whose input is a combination (concatenation, addition or both) of $t$ and $\vx$.
Hence we can write $h_\theta (t \mid \vx)$ and $h_\theta (t, \vx)$ interchangeably.
The gradient $\nabla_\vx h_\theta (t, \vx)$ is naturally defined and so is $\nabla_vx h_\theta (t \mid \vx)$.
}

\begin{equation} \label{eq:hazard_gradient_penalty}
\mathcal{R}_\vx
= \E_{t \sim S_\theta(t \mid \vx)}
\left[
    \left\Vert \nabla_\vx h_\theta(t \mid \vx) \right\Vert_2
\right]
\end{equation}

\subsection{Efficient Sampling from the Survival Density}

The sampling operation $t \sim S(t \mid \vx)$\footnote{
$S(t \mid \vx)$ is not a valid probability distribution as we cannot guarantee $\int S(t \mid \vx) dt = 1$.
Rigorously, we sample $t \sim s(t \mid \vx)$ where $s(t \mid \vx) = S(t \mid \vx) / \int S(t \mid \vx) dt$.
We use $t \sim S(t \mid \vx)$ for notational simplicity.
} in \eqref{eq:hazard_gradient_penalty} may induce computational overhead.
To boost the sampling operation, we use $\log S_\theta(t \mid \vx)$ which was computed during the negative log-likelihood calculation in \eqref{eq:negative_log_likelihood}.
Let $[t_1, \dots, t_K]$ be the union of the time points in minibatch.
The time points are sorted in increasing order.
The adaptive time stepping in ODE solvers are sensitive to the time interval $t_K - t_1$ rather than the number of time points \cite{rubanova2019latent}.
We can access $\log S_\theta(t_k \mid \vx)$ with negligible overhead as long as $t_1 < t_k < t_K$.

We sample $t_k$ from a categorical distribution whose $k$-th weight is defined as $S_\theta(t_k \mid \vx) = \exp( - \int_0^t h(\tau \mid \vx) d\tau )$.
We finalize the sampling process by sampling $t$ from the uniform distribution $\mathcal{U}([t_k, t_{k+1}])$.
In this way, we don't have to calculate $S(t \mid \vx)$ again for sampling $t$.

\begin{algorithm}
\caption{Hazard Gradient Penalty}
\label{alg:hazard_gradient_penalty}
\begin{algorithmic}
\Require $h_\theta$, learning rate $\gamma$
\Repeat
    \State sample $(\vx, t, e) \sim \mathcal{D}$
    \State Retrieve unique times $t_1, \dots, t_K$ from minibatch.
    \State integrate $-h_\theta(t \mid \vx)$ from 0 to $t_K$ and store $\log S_\theta(t_1 \mid \vx), \dots, \log S_\theta(t_K \mid \vx)$
    \State $\log S_\theta(t \mid \vx) \gets$ choose from $\log S_\theta(t_1 \mid \vx), \dots, \log S_\theta(t_K \mid \vx)$ that corresponds to $(\vx, t, e)$
    \State $\log p_\theta(t \mid \vx) \gets \log h_\theta (t \mid \vx) + \log S_\theta (t \mid \vx)$
    \State $\mathcal{L}_\vx \gets - e \log p_\theta(t \mid \vx) + (1 - e) \log S_\theta(t \mid \vx)$ \Comment{Negative log-likelihood}
    \State sample $i_1, \dots, i_M \sim \mathrm{Categorical}(S_\theta(t_1 \mid \vx), \dots, S_\theta(t_K \mid \vx))$ \Comment{$t \sim S_\theta(t \mid \vx)$}
    \State $t^\prime_m \gets t_{i_m - 1} + \mathrm{Uniform}(t_{i_m - 1}, t_{i_m})$ \Comment{$t_0 = 0$}
    \State $\mathcal{R}_\vx \gets \frac{1}{M} \sum_{m=1}^M\Vert \nabla_\vx h_\theta (t^\prime_m \mid \vx) \Vert_2$ \Comment{Hazard gradient penalty}
    \State $\theta \gets \theta - \gamma \nabla_\theta (\mathcal{L}_\vx + \lambda \mathcal{R}_\vx)$
\Until{Convergence}
\end{algorithmic}
\end{algorithm}

\subsection{Connection to KL Divergence} \label{sec:kl_div_connection}

% The KL divergence between two distributions $p(t), q(t)$ is defined as the expected difference between the log-densities:

% \begin{equation}
% \E_{p(t)} \left[ \log p(t) - \log q(t) \right]
% \end{equation}

We now show that the hazard gradient penalty in \eqref{eq:hazard_gradient_penalty} is related to minimizing the KL divergence between the density function at a data point and that of the neighborhood points of the data point.

\begin{theorem} \label{thm:kl_upper_bound}
Suppose the hazard function is strictly positive function for all data point $\vx$.
The KL divergence
\begin{equation} \nonumber
\E_{p(t \mid \vx)} \left[ \log p(t \mid \vx) - \log p(t \mid \vx^\prime) \right]
\end{equation}
is upper bounded by 
\begin{equation} \label{eq:kl_upper_bound}
\E_{p(t \mid \vx)} \left\Vert \log h(t \mid \vx) - \log h(t \mid \vx^\prime) \right\Vert_2
+ \E_{S(t \mid \vx)} \left\Vert h(t \mid \vx) - h(t \mid \vx^\prime) \right\Vert_2
\end{equation}
\end{theorem}

To prove Theorem \ref{thm:kl_upper_bound}, we need the following  lemma.

\begin{lemma} \label{thm:lemma}
The expectation of survival densities difference under the density is
the negative of the expectation of hazard functions difference under the survival density.
In other words,
\begin{equation*}
\E_{p(t \mid \vx)} \left[ \log S(t \mid \vx) - \log S(t \mid \vx^\prime) \right]
= - \E_{S(t \mid \vx)} \left[ h(t \mid \vx) - h(t \mid \vx^\prime) \right]
\end{equation*}
\end{lemma}

Proof) We use the fact that $\E_{S(t \mid \vx)} \left( \log S(t \mid \vx) - \log S(t \mid \vx^\prime) \right)$ is constant with respect to $t$.
\begin{align*}
&\frac{d}{dt} \E_{S(t \mid \vx)} \left( \log S(t \mid \vx) - \log S(t \mid \vx^\prime) \right) \\
&= \frac{d}{dt} \sum S(t \mid \vx) \left( \log S(t \mid \vx) - \log S(t \mid \vx^\prime) \right) \\
&= - \sum p(t \mid \vx) \left( \log S(t \mid \vx) - \log S(t \mid \vx^\prime) \right)
 + \sum S(t \mid \vx) \left( - h(t \mid \vx) + h(t \mid \vx^\prime) \right) = 0
\end{align*}

Hence,
\begin{equation*}
\E_{p(t \mid \vx)} \left[ \log S(t \mid \vx) - \log S(t \mid \vx^\prime) \right]
= - \E_{S(t \mid \vx)} \left[ h(t \mid \vx) - h(t \mid \vx^\prime) \right] \blacksquare
\end{equation*}

We now go back to Theorem \ref{thm:kl_upper_bound} and prove the theorem.
\begin{align*}
&\E_{p(t \mid \vx)} \left[ \log p(t \mid \vx) - \log p(t \mid \vx^\prime) \right] \\
&= \left\Vert \E_{p(t \mid \vx)} \left[ \log p(t \mid \vx) - \log p(t \mid \vx^\prime) \right] \right\Vert_2 \; (\because D_{KL} \geq 0) \\
&= \big\Vert
\E_{p(t \mid \vx)} \left[ \log h(t \mid \vx) - \log h(t \mid \vx^\prime) \right]
 - \E_{p(t \mid \vx)} \left[ \log S(t \mid \vx) - \log S(t \mid \vx^\prime) \right]
\big\Vert_2 \; (\because \text{\eqref{eq:hazard_func_def}})\\
&= \big\Vert
\E_{p(t \mid \vx)} \left[ \log h(t \mid \vx) - \log h(t \mid \vx^\prime) \right]
 + \E_{S(t \mid \vx)} \left[ h(t \mid \vx) - h(t \mid \vx^\prime) \right]
\big\Vert_2 \; (\because \text{Lemma } \ref{thm:lemma})\\
&\leq \left\Vert
    \E_{p(t \mid \vx)} \left[ \log h(t \mid \vx) - \log h(t \mid \vx^\prime) \right]   
\right\Vert_2
 + \left\Vert
    \E_{S(t \mid \vx)} \left[ h(t \mid \vx) - h(t \mid \vx^\prime) \right]
\right\Vert_2 (\because \text{triangle inequality}) \\
&\leq 
    \E_{p(t \mid \vx)} \left\Vert \log h(t \mid \vx) - \log h(t \mid \vx^\prime) \right\Vert_2
 + 
    \E_{S(t \mid \vx)} \left\Vert h(t \mid \vx) - h(t \mid \vx^\prime) \right\Vert_2 \blacksquare
 \\
\end{align*}

% fill in theorem proof
Now suppose that $\vx^\prime$ is in $\epsilon$-ball centered at $\vx$.
We show that the approximation acquired by Taylor approximation
$h(t \mid \vx^\prime) \approx h(t \mid \vx) + \nabla_\vx h(t \mid \vx)^T (\vx^\prime - \vx)$ is upper bounded by
$2 \epsilon \E_{S(t \mid \vx)} \left\Vert \nabla_\vx h(t \mid \vx) \right\Vert_2$
% \begin{equation} \label{eq:kl_upper_bound_bound}
% 2 \epsilon \E_{S(t \mid \vx)} \left\Vert \nabla_\vx h(t \mid \vx) \right\Vert_2
% \end{equation}

The second term in \eqref{eq:kl_upper_bound} is approximated as follows.
\begin{equation} \label{eq:hazard_diff_approx}
\E_{S(t \mid \vx)} \left\Vert h(t \mid \vx) - h(t \mid \vx^\prime) \right\Vert_2
\approx \E_{S(t \mid \vx)} \left\Vert \nabla_\vx h(t \mid \vx)^T (\vx^\prime - \vx) \right\Vert_2
\end{equation}
which is equal to \eqref{eq:hazard_gradient_penalty} up to constant multiplication.

As $\vx^\prime \in \mathcal{B}(\vx, \epsilon)$, the maximum of \eqref{eq:hazard_diff_approx} is achieved when
$\vx^\prime - \vx = \epsilon \nabla_\vx h(t \mid \vx)$.
% \begin{equation} \nonumber
% \vx^\prime - \vx = \epsilon \nabla_\vx h(t \mid \vx)
% \end{equation}
Hence the maximum of \eqref{eq:hazard_diff_approx} is
$\epsilon \E_{S(t \mid \vx)} \left\Vert \nabla_\vx h(t \mid \vx) \right\Vert_2$.

Similarly, the first term in \eqref{eq:kl_upper_bound} is approximated as follows.
\begin{align*}
\E_{p(t \mid \vx)} \left\Vert \log h(t \mid \vx) - \log h(t \mid \vx^\prime) \right\Vert_2
&\approx \E_{p(t \mid \vx)} \left\Vert \nabla_\vx \log h(t \mid \vx)^T (\vx^\prime - \vx) \right\Vert_2 \\
&= \E_{p(t \mid \vx)} \left\Vert \frac{\nabla_\vx h(t \mid \vx)^T}{h(t \mid \vx)} (\vx^\prime - \vx) \right\Vert_2 \\
&= \sum \frac{p(t \mid \vx)}{h(t \mid \vx)} \left\Vert \nabla_\vx h(t \mid \vx)^T (\vx^\prime - \vx) \right\Vert_2 (\because h(\cdot \mid \vx) \geq 0) \\
&= \E_{S(t \mid \vx)} \left\Vert \nabla_\vx h(t \mid \vx)^T (\vx^\prime - \vx) \right\Vert_2 \; (\because \eqref{eq:hazard_func_def})
\end{align*}

The maximum of the approximation $\E_{p(t \mid \vx)} \left\Vert \nabla_\vx \log h(t \mid \vx)^T (\vx^\prime - \vx) \right\Vert_2$ is also 
$\epsilon \E_{S(t \mid \vx)} \left\Vert \nabla_\vx h(t \mid \vx) \right\Vert_2$.
 Hence, under the assumption that $\vx^\prime \in B(\vx, \epsilon)$, the approximation of the \eqref{eq:kl_upper_bound} is bounded by the constant multiplication of \eqref{eq:hazard_gradient_penalty}.

To incorporate the regularizer into the negative log-likelihood loss, we minimize the Lagrange multiplier defined as the sum of the negative log-likelihood and the hazard gradient penalty regularizer.
\begin{equation} \label{eq:hgp_loss}
\mathcal{L} = \E_{(\vx, t, e) \sim \mathcal{D}} \left[ \mathcal{L}_\vx + \lambda \mathcal{R}_\vx \right]
\end{equation}
Here, $\lambda$ is a coefficient that balances the negative log-likelihood and the regularizer.

Minimizing the hazard gradient penalty in \eqref{eq:hazard_gradient_penalty} has two advantages over minimizing the KL divergence directly: a) computational efficiency and b) reduced burden of hyperparameter tuning.
To compute the KL divergence, we first sample $\vx^\prime \in B(\vx, \epsilon)$.
We then need to compute four values: $h(t | \vx), S(t | \vx), h(t | \vx^\prime)$ and $S(t | \vx^\prime)$.
In this case, we have to compute hazard values of every $t \sim S(t | \vx)$.
Further, we need one more hazard function integration $S(t \mid \vx^\prime) = \exp (- \int h(t \mid \vx^\prime))$.
On the other hand, regularizing the hazard gradient penalty only need to calculate the gradient of the hazard function.

When it comes to regularizing the KL divergence, we have to set the appropriate value of the regularizing coefficient $\lambda^\prime$ and the size of the ball $\epsilon$.
On the other hand, if we regularize the hazard gradient penalty, we don’t need to tune $\epsilon$ as $\lambda$ in \eqref{eq:hazard_gradient_penalty} incorporates $\epsilon$.

\subsection{The Source of Performance Enhancement}

It is hard to expect that minimizing the KL divergence between the density function at a data point and that of its neighbours will enhance the model’s performance.
However, once we accept the cluster assumption, we can expect that the regularizer in \eqref{eq:hazard_gradient_penalty} will improve the performance of the model.

Consider a case where two data points $\vx_1, \vx_2$ from the training set failed at $t_1, t_2 (t_1 < t_2)$ each.
Under the cluster assumption, a point $\vx_2^\prime \in B(\vx_2, \epsilon)$ in the test set should fail at $t_2^\prime \approx t_2$.
If $\hat{p}(t \mid \vx_2^\prime)$ is skewed for some reason and puts high density at $t_1^\prime < t_1$, it worsens the model’s performance.
To evade such situation,  $\hat{p}(t \mid \vx_2^\prime)$ should not deviate too much from $\hat{p}(t \mid \vx_2)$.
Our regularizer solves this problem by minimizing the KL divergence between the density function at a data point and that of its neighbours.

\section{Experiments}

In this section, we experimentally show that the proposed method outperforms other regularizers.
Further, we check the hyperparameter sensitivity of our proposed method.
Throughout the experiments, we use three public datasets: Study to Understand Prognoses Preferences Outcomes and Risks of Treatment (SUPPORT) \footnote{https://github.com/autonlab/auton-survival/blob/master/dsm/datasets/support2.csv},
the Molecular Taxonomy of Breast Cancer International Consortium (METABRIC) \footnote{https://github.com/jaredleekatzman/DeepSurv/tree/master/experiments/data/metabric},
and the Rotterdam tumor bank and German Breast Cancer Study Group (RotGBSG) \footnote{https://github.com/jaredleekatzman/DeepSurv/tree/master/experiments/data/gbsg}. 
Table \ref{tab:dataset_details} summarizes the statistics of the datasets.

\begin{table*}
  \caption{
      Summary statistics of the datasets used in our experiments.
      $N$ denotes the number of data points and $d$ denotes the dimension of each data points.
  }
  \label{tab:dataset_details}
  \setlength\tabcolsep{4pt}
  \centering
  \begin{tabular}{l|r|r|r|r|r|r|r|r}
    \toprule
    \multirow{2}{*}{Dataset}& \multirow{2}{*}{$N$} &\multirow{2}{*}{$d$} & \multirow{2}{*}{Censoring ($\%$)} & \multicolumn{2}{c|}{Durations} & \multicolumn{3}{c}{Event Quantiles}  \\
    \cline{5-9}
    & & & & \# unique & domain & $t=25$\% & $t=50$\% & $t=75$\%  \\
    \midrule
    SUPPORT & 9105 & 43 & 31.89\% & 1724 & $\mathbb{N}^+$ & 14 & 58 & 252 \\
    METABRIC & 1904 & 9 & 42.06\% & 1686 & $\mathbb{R}^+$ & 42.68 & 85.86 & 145.33 \\
    RotGBSG & 2232 & 7 & 43.23\% & 1230 & $\mathbb{R}^+$ & 13.61 & 24.01 & 40.32 \\
    % WHAS & 500 & 14 & 57.00\% & 395 & $\mathbb{N}^+$ & 21 & 166 & 613 \\
  \bottomrule
\end{tabular}
\end{table*}

\subsection{Evaluation Metrics}

Throughout this subsection,
we denote $\hat{S}(t \mid \vx)$ as the estimate of $S(t \mid \vx)$,
$I(\cdot)$ as the indicator function,
$(\vx_i, T_i, e_i)$ as the $i$th covariate, time, event indicator of the dataset,
$\hat{G}(t)$ as the Kaplan-Meier estimator for censoring distribution \citep{kaplan1958nonparametric},
and $\omega_i$ as $1/\hat{G}(T_i)$.

\subsubsection{Time Dependent Concordance Index ($C^{td}$)} \label{sec:c_td}

The concordance index, or C-index is defined as the
proportion of correctly ordered pairs among all comparable pairs. We use time dependent variant of
C-index that truncates pairs within the prespecified time point \cite{uno2011cindex}.
The time dependent concordance index at $t$, $C^{td}(t)$, is defined as 
\begin{equation} \nonumber
\frac
{\sum_{i=1}^N \sum_{j=1}^N e_i \{ \hat{G}(T_i) \}^{-2} I(T_i < T_j, T_i < t) I(\hat{S}(t \mid \vx_i) < \hat{S}(t \mid \vx_j))}
{\sum_{i=1}^N \sum_{j=1}^N e_i \{ \hat{G}(T_i) \}^{-2} I(T_i < T_j, T_i < t)}
\end{equation}

To evaluate $C^{td}$ at $[t_1, \dots, t_L]$ at the same time, we take its mean
$mC^{td} = \frac{1}{L} \sum_{l = 1}^L C^{td}(t_l)$.

\subsubsection{Time Dependent Area Under Curve (AUC)}
is an extension of the ROC-AUC to survival data \cite{hung2010auc}.
It measures how well a model can distinguish individuals who fail before the given time ($T_i < t$) and who fail after the given time ($T_j > t$).

The AUC at time $t$, $AUC(t)$, is defined as 
\begin{equation} \nonumber
\frac{\sum_{i=1}^N \sum_{j=1}^N I(T_j > t) I(T_i \leq t) \omega_i I(\hat{S}(t \mid \vx_i) \leq \hat{S}(t \mid \vx_j))}
{(\sum_{i=1}^N I(T_i > t))(\sum_{i=1}^N I(T_i \leq t) \omega_i)}
\end{equation}

To evaluate $AUC$ at $[t_1, \dots, t_L]$ at the same time, we take its mean $mAUC = \frac{1}{L} \sum_{l=1}^L AUC(t_l)$.

\subsubsection{Negative Binomial Log-Likelihood}

We can evaluate the negative binomial log-likelihood (NBLL) to measure both discrimination and calibration performance \cite{kvamme2019time}.
The negative binomial log-likelihood at $t$ measures how close the survival probability is to 1 if the given data survived at $t$ and how close the survival probability is to 0 if the given data failed before $t$.
The NBLL at $t$, $NBLL(t)$, is defined as 
\begin{equation} \nonumber
- \frac{1}{N} \sum_{i=1}^N
\left[
    \frac{\log(1 - \hat{S}(t \mid \vx_i)) I ( T_i \leq t, e_i = 1 )}{\hat{G}(T_i)} 
    + \frac{\log \hat{S}(t \mid \vx_i) I(T_i > t)}{\hat{G}(t)}
\right]
\end{equation}

For the convenience of evaluation, we integrate the NBLL, 
$iNBLL = \frac{1}{t_2 - t_1} \int_{t_1}^{t_2} NBLL(t) dt$.

\subsection{Methods Compared}

We compare our proposed method with four methods: \textit{vanilla ODE}, \textit{ODE + L1}, \textit{ODE + L2} \textit{ODE + LCI}.
\textit{Vanilla ODE} minimizes the expectation of the negative log-likelihood in \eqref{eq:negative_log_likelihood}.
\textit{ODE + L1} minimizes the Lagrange multiplier defined as the sum of the expectation of the negative log-likelihood and the L1 penalty term.
\begin{equation} \nonumber
\E_{(\vx, t, e) \sim \mathcal{D}} \mathcal{L}_\vx + \alpha \sum_{p=1}^P |w_p|
\end{equation}
Here, $w_p$s are model parameters and $\alpha$ is a coefficient that balances the negative log-likelihood and the L1 penalty term.
This is an extension of Lasso-Cox \cite{tibshirani1997lasso} to the ODE modeling framework.
\textit{ODE + L2} minimizes the Lagrange multiplier defined as the sum of the expectation of the negative log-likelihood and the L2 penalty term.
\begin{equation} \nonumber
\E_{(\vx, t, e) \sim \mathcal{D}} \mathcal{L}_\vx + \alpha \sum_{p=1}^P w_p^2
\end{equation}
Here, $w_p$s are model parameters and $\alpha$ is a coefficient that balances the negative log-likelihood and the L2 penalty term.
This is an extension of Ridge-Cox \cite{verweij1994penalized} to the ODE modeling framework.
\textit{ODE + LCI} minimizes the Lagrange multiplier defined as the sum of the expectation of the negative log-likelihood and the negative of the lower bound of a simplified version of time-dependent C-index.
The regularizer is defined as
\begin{equation} \nonumber
- \sum_t \frac{
    \sum_{i=1}^N \sum_{j=1}^N e_i I(T_i < T_j, T_i < t) (1 + (\log \sigma(S_\theta (t \mid \vx_i) < S_\theta (t \mid \vx_j)) / \log 2)
}
{\sum_{i=1}^N \sum_{j=1}^N e_i I(T_i < T_j, T_i < t)}
\end{equation}
This is equivalent to time dependent concordance index in Section \ref{sec:c_td} if we don't take the Kaplan-Meier estimator into account.
The regularizer is a reminiscent of the lower bound of C-index \cite{steck2007bound}.
Although the lower bound of C-index was originally proposed as a substitute of the negative log-likelihood, \citet{chapfuwa18a2018adversarial} used the lower bound \cite{steck2007bound} as a regularizer of the AFT model \cite{wei1992accelerated}.

\subsection{Experimental Details} \label{sec:experimental_details}

Across all datasets, we split training set, validation set and test set into 70\%, 10\% and 20\% each using \texttt{PyTorch}’s \texttt{random\_split} function \cite{paszke2019pytorch}.
We set \texttt{seed = 42} when splitting.

Across all experiments, we use an MLP with two hidden layers where each layer has 64 hidden units.
Across all layers, we apply Layer normalization \cite{ba2016layernorm}.
Instead of naively feeding time $t$ into the neural network, we feed scaled time $\tilde{t} = (t - t_2) / (t_3 - t_1)$ where $t_1, t_2$, and $t_3$ are first, second, and third quartile of failure event times.
We found that this strategy enhances the ODE model's performance and boosts training time.
To incorporate time $t$ into the survival analysis model, we project the time into an eight dimensional vector using a single layer MLP and then concatenate it to the input data.
% We also add projected time information with output layerwise.
The time $t$ is also specified by adding projected output into each layer output.
We use the AdamW optimizer \cite{loshchilov2018decoupled} and clipped the gradient norm so that it does not exceed 1.
We set the learning rate to 0.001.
We have implemented the code using \texttt{JAX} \cite{jax2018github} and \texttt{Diffrax} \cite{kidger2021on} \footnote{The code will be made publicly available in the near future.}.

To find the best $\lambda$ in \eqref{eq:hgp_loss}, we run experiments with $\lambda = 1, 5, 10, 50$ and report the results at $\lambda = 10$ as it shows decent performance across all metrics and datasets.
We also have to set the number of samples $M$ from the time sampling process $t \sim p_\theta(t \mid \vx)$ in \eqref{eq:hazard_gradient_penalty}.
We set $M = 5$ across all the hazard gradient penalty experiments.
To find the best coefficient $\alpha$ in \textit{ODE + L1}, \textit{ODE + L2}, and \textit{ODE + LCI} experiments, we set $\alpha = 1e-1, 1e-2, 1e-3$ and run the experiments.
We report the best $\alpha$ in terms of $mAUC$.
To report $mC^{td}$ and $mAUC$, we calculate $C^{td}$ and $AUC$ at 10\%, 20\%, $\dots$, 90\% event quantiles and average them.
To report $iNBLL$, we integrate from the minimum time of the test set to the maximum time of the test set.
We use \texttt{scikit-survival} \cite{polsterl2020sksurv} to report $mC^{td}$ and $mAUC$.
We use \texttt{pycox} \cite{kvamme2019time} to report $iNBLL$.
Across all experiments, we run 5 experiments with different seeds and report their mean and the standard deviation.

\subsection{Results}

\begin{table}[t]
     \caption{Experimental Results on three datasets.}
     \label{tab:experimental_result}
    \begin{subtable}[h]{\textwidth}
        \caption{$mC^{td} (\uparrow)$}
        \label{tab:c_td_results}
        \centering
        \setlength\tabcolsep{4pt}
        \begin{tabular}{l||c|c|c}
        \toprule
        Method & SUPPORT & METABRIC & RotGBSG \\
        \midrule
        ODE & 0.771 $\pm$ 0.003 & 0.695 $\pm$ 0.009 & 0.718 $\pm$ 0.004 \\
        ODE + L1 & 0.771 $\pm$ 0.003 & 0.696 $\pm$ 0.008 & 0.718 $\pm$ 0.003 \\
        ODE + L2 & 0.772 $\pm$ 0.002 & 0.695 $\pm$ 0.007 & 0.718 $\pm$ 0.003 \\
        ODE + LCI & 0.771 $\pm$ 0.003 & \textbf{0.700 $\pm$ 0.001} & 0.716 $\pm$ 0.004 \\
        \midrule
        ODE + HGP & \textbf{0.776 $\pm$ 0.002} & \textbf{0.701 $\pm$ 0.008} & \textbf{0.721 $\pm$ 0.005} \\
        \end{tabular}
    \end{subtable}
    \vfill
    \begin{subtable}[h]{\textwidth}
        \caption{$mAUC (\uparrow)$}
        \label{tab:auc_results}
        \centering
        \setlength\tabcolsep{4pt}
        \begin{tabular}{l||c|c|c}
        \toprule
        Method & SUPPORT & METABRIC & RotGBSG \\
        \midrule
        ODE & 0.809 $\pm$ 0.002 & 0.730 $\pm$ 0.007 & 0.745 $\pm$ 0.003 \\
        ODE + L1 & 0.809 $\pm$ 0.001 & 0.730 $\pm$ 0.007 & 0.745 $\pm$ 0.003 \\
        ODE + L2 & 0.810 $\pm$ 0.001 & 0.730 $\pm$ 0.006 & 0.745 $\pm$ 0.003 \\
        ODE + LCI & 0.809 $\pm$ 0.002 & \textbf{0.733 $\pm$ 0.002} & 0.744 $\pm$ 0.003 \\
        \midrule
        ODE + HGP & \textbf{0.814 $\pm$ 0.001} & \textbf{0.733 $\pm$ 0.006} & \textbf{0.751 $\pm$ 0.004} \\
        \end{tabular}
     \end{subtable}
     \vfill
    \begin{subtable}[h]{\textwidth}
        \caption{$iNBLL (\downarrow)$}
        \label{tab:nbll_results}
        \centering
        \setlength\tabcolsep{4pt}
        \begin{tabular}{l||c|c|c}
        \toprule
        Method & SUPPORT & METABRIC & RotGBSG \\
        \midrule
        ODE & 0.518 $\pm$ 0.016 & 0.474 $\pm$ 0.005 & 0.530 $\pm$ 0.008 \\
        ODE + L1 & 0.517 $\pm$ 0.016 & 0.473 $\pm$ 0.005 & 0.534 $\pm$ 0.010 \\
        ODE + L2 & 0.515 $\pm$ 0.014 & 0.472 $\pm$ 0.003 & 0.534 $\pm$ 0.010 \\
        ODE + LCI & 0.518 $\pm$ 0.016 & \textbf{0.469 $\pm$ 0.002} & \textbf{0.527 $\pm$ 0.009} \\
        \midrule
        ODE + HGP & \textbf{0.504 $\pm$ 0.011} & 0.473 $\pm$ 0.004 & \textbf{0.528 $\pm$ 0.002} \\
        \end{tabular}
     \end{subtable}
\end{table}

Table \ref{tab:experimental_result} shows the $mC^{td}, mAUC$, and $iNBLL$ scores.
The hazard gradient penalty outperforms other methods across almost all metrics and datasets.
The interesting point is that both L1 and L2 penalties do not affect the ODE model's performance in most cases.
We speculate that regularizing the weight norm is effective in CoxPH as the model is simple and has a strong assumption that the hazard rate is constant.
On the contrary, regularizing the norm of the weight may not be able to affect the ODE model's performance as ODE models are much more complex than CoxPH.
Also, the experimental results highlight the possibility that the performance of the survival analysis models is more related to the local information such as the gradient at each data point rather than the global information such as the weight norm of the model.

Table \ref{tab:experimental_result} also shows that regularizing the lower bound of the C-index is not effective in many cases.
% Though the method performs well on the METABRIC dataset, it affects little to none on other datasets.
We conjecture that the method is ineffective as the ODE modeling framework is flexible and optimizing the negative log-likelihood can discriminate each data point’s rank.
Furthermore, regularizing the lower bound of the C-index does not harness the information of neighbors of data points.
This information gap leads to the performance gap between \textit{ODE + HGP} and \textit{ODE + LCI}.

\begin{table}
  \caption{Experimental Results SUPPORT ablated in terms of sample size $M$. We set $\lambda = 10$.}
  \label{tab:varying_num_samples}
  \centering
  \setlength\tabcolsep{4pt}
  \begin{tabular}{l||c|c|c}
  \toprule
  Method & $mC^{td} (\uparrow)$ & $mAUC (\uparrow)$ & $iNBLL (\downarrow)$ \\
  \midrule
  No reg. & 0.771 $\pm$ 0.003 & 0.809 $\pm$ 0.002 & 0.518 $\pm$ 0.016 \\
  $M = 1$ & \textbf{0.776 $\pm$ 0.003} & \textbf{0.814 $\pm$ 0.001} & \textbf{0.503 $\pm$ 0.010} \\
  $M = 5$ & \textbf{0.776 $\pm$ 0.002} & \textbf{0.814 $\pm$ 0.001} & \textbf{0.504 $\pm$ 0.011} \\
  $M = 10$ & \textbf{0.776 $\pm$ 0.002} & \textbf{0.815 $\pm$ 0.001} & \textbf{0.503 $\pm$ 0.009} \\
  \end{tabular}
\end{table}

\begin{figure*}
     \centering
     \begin{subfigure}[b]{0.32\textwidth}
         \centering
         \includegraphics[width=\textwidth]{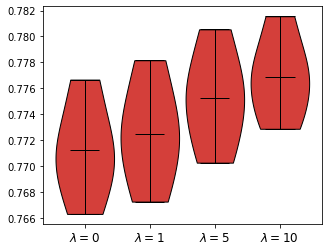}
         \caption{SUPPORT $mC^{td} (\uparrow)$}
         \label{fig:violin_SUPPORT_c_td}
     \end{subfigure}
     \hfill
     \begin{subfigure}[b]{0.32\textwidth}
         \centering
         \includegraphics[width=\textwidth]{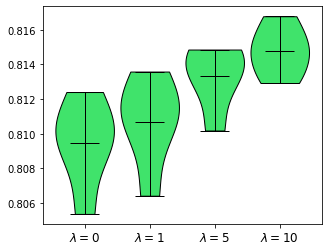}
         \caption{SUPPORT $mAUC (\uparrow)$}
         \label{fig:violin_SUPPORT_auc}
     \end{subfigure}
     \hfill
     \begin{subfigure}[b]{0.32\textwidth}
         \centering
         \includegraphics[width=\textwidth]{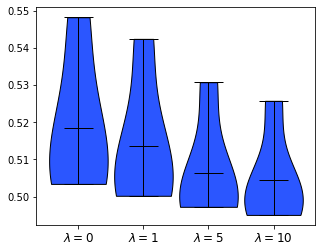}
         \caption{SUPPORT $iNBLL (\downarrow)$}
         \label{fig:violin_SUPPORT_nbll}
     \end{subfigure}
     \vfill
     \begin{subfigure}[b]{0.32\textwidth}
         \centering
         \includegraphics[width=\textwidth]{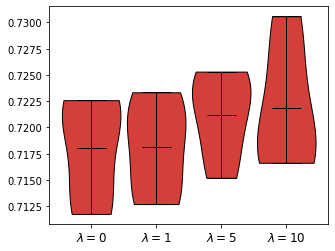}
         \caption{RotGBSG $mC^{td} (\uparrow)$}
         \label{fig:violin_RotGBSG_c_td}
     \end{subfigure}
     \hfill
     \begin{subfigure}[b]{0.32\textwidth}
         \centering
         \includegraphics[width=\textwidth]{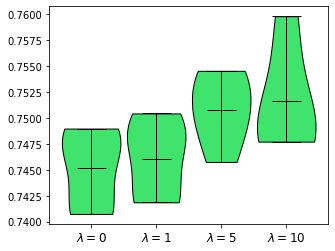}
         \caption{RotGBSG $mAUC (\uparrow)$}
         \label{fig:violin_RotGBSG_auc}
     \end{subfigure}
     \hfill
     \begin{subfigure}[b]{0.32\textwidth}
         \centering
         \includegraphics[width=\textwidth]{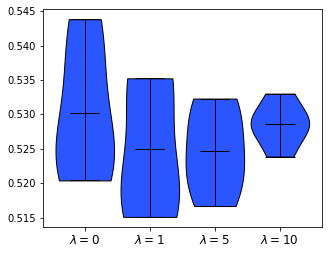}
         \caption{RotGBSG $iNBLL (\downarrow)$}
         \label{fig:violin_RotGBSG_nbll}
     \end{subfigure}
    \caption{
    Violin plots of experimental results on SUPPORT and RotGBSG by varying $\lambda$.
    \textcolor{red}{Red}, \textcolor{green}{green}, and \textcolor{blue}{blue} denote \textcolor{red}{$mCtd$}, \textcolor{green}{$mAUC$}, and \textcolor{blue}{$iNBLL$}.
    The thickness of a plot denotes the probability density of the results.
    % We don’t show the results when $\lambda > 10$ as the performance degraded as $\lambda$ increases in those cases.
    % The non-linear trend of RotGBSG $iNBLL$ shows that a high $mC^{td}$ does not always accompany a low $iNBLL$.
    Our proposed regularizer may conflict with the negative log-likelihood if we set high $\lambda$.
    The $\lambda$ that achieves the best scores across all metrics on the RotGBSG dataset could have been acquired between $\lambda = 5$ and $\lambda = 10$.
    However, we report the result at $\lambda = 5$ on the RotGBSG dataset in Table \ref{tab:nbll_results} for consistency.
    }
    \label{fig:varying_lambda_violin_plots}
\end{figure*}

Table \ref{tab:varying_num_samples} shows the results by varying the number of samples $M$ in the sampling process $t \sim S(t \mid \vx)$ in \eqref{eq:hazard_gradient_penalty}.
As long as the regularizer is applied, the number of samples $M$ does not affect the performance.
Even when $M = 1$, the regularizer works well.
Figure \ref{fig:varying_lambda_violin_plots} shows the results on SUPPORT and RotGBSG datasets by varying the coefficient $\lambda$ in \eqref{eq:hgp_loss}.
Since the performance variation by $\lambda$ is stable, the hyperparameter $\lambda$ can be tuned without much difficulty in practical setups.

\section{Related Works}

A line of research integrated deep neural networks to CoxPH \cite{faraggi1995neural, katzman2018deepsurv} and Extended Hazards \cite{zhong2022deepeh} for more model flexibility.
Another line of research proposed distribution-free survival analysis models via the time domain discretization \cite{lee2018deephit} or adversarial learning approach \cite{chapfuwa18a2018adversarial}.
Previous works \cite{goldstein2020xcal, han2021survivalgames} proposed new objectives to optimize Brier score \cite{graf1999assessment}, Binomial log-likelihood, or distributional calibration directly.
Yet to the best of our knowledge, none of the previous works focused on the effect of gradient penalty on survival analysis models.

Previous works proposed L1 and L2 regularization in the survival analysis literature \cite{tibshirani1997lasso, verweij1994penalized}.
Those methods regularize the survival analysis models so that the L1 or L2 norm of the model parameters does not increase so much.
Our method is different from those methods in that we penalize the norm of the gradient on each local data point.

Our method is closely related to semi-supervised learning \cite{chapelle2006semi}.
Among many semi-supervised learning methods, our method is germane to virtual adversarial training \cite{miyato2018virtual} in that it regularizes function variation between a local data point and its neighbours.
However, virtual adversarial training is different from ours in that the method was demonstrated in the classification setting and the output is a discrete distribution.

In Generative Adversarial Nets (GANs) literature \cite{goodfellow2014gan}, the gradient penalty had been studied actively.
\citet{gulrajani2017wgangp} proposed the gradient penalty to satisfy the 1-Lipschitz function constraint in Kantrovich-Rubinstein duality.
\citet{mescheder2018gantraining} proposed the gradient penalty to penalize the discriminator for deviating from the Nash equilibrium.
Ours is different from these works in that we propose gradient penalty so that the density at $\vx$ does not deviate much from that of $\vx$’s neighborhood points.

\section{Conclusion} \label{sec:conclusion}

In this paper, we introduced a novel regularizer for survival analysis.
Unlike previous methods, we focus on individual local data point rather than global information.
We theoretically showed that regularizing the norm of the gradient of hazard function with respect to the data point is related to minimizing the KL divergence between the data point and that of its neighbours.
Empirically, we showed that the proposed regularizer outperforms other regularizers and it is not sensitive to hyperparameters.
Nonetheless, as minimizing the proposed regularizer may conflict with optimizing the negative log-likelihood, practitioners should tune the balancing coefficient $\lambda$ for each dataset.
The paper highlights the new possibility that the recent advancements in semi-supervised learning could enhance the performance of survival analysis models.

\paragraph{Social Impact}
Survival analysis models are widely used in health care, economics, engineering, and business.
We observed that our proposed method enhances the survival analysis model’s performance.
Nevertheless, our method may not be the silver bullet and should be treated with care.

\clearpage

\bibliographystyle{abbrvnat}
\bibliography{ref}

%%%%%%%%%%%%%%%%%%%%%%%%%%%%%%%%%%%%%%%%%%%%%%%%%%%%%%%%%%%%
\clearpage

\section*{Checklist}

%%% BEGIN INSTRUCTIONS %%%
% The checklist follows the references.  Please
% read the checklist guidelines carefully for information on how to answer these
% questions.  For each question, change the default \answerTODO{} to \answerYes{},
% \answerNo{}, or \answerNA{}.  You are strongly encouraged to include a {\bf
% justification to your answer}, either by referencing the appropriate section of
% your paper or providing a brief inline description.  For example:
% \begin{itemize}
%   \item Did you include the license to the code and datasets? \answerYes{See Section~\ref{gen_inst}.}
%   \item Did you include the license to the code and datasets? \answerNo{The code and the data are proprietary.}
%   \item Did you include the license to the code and datasets? \answerNA{}
% \end{itemize}
% Please do not modify the questions and only use the provided macros for your
% answers.  Note that the Checklist section does not count towards the page
% limit.  In your paper, please delete this instructions block and only keep the
% Checklist section heading above along with the questions/answers below.
%%% END INSTRUCTIONS %%%

\begin{enumerate}

\item For all authors...
\begin{enumerate}
  \item Do the main claims made in the abstract and introduction accurately reflect the paper's contributions and scope?
    \answerYes{}
  \item Did you describe the limitations of your work?
    \answerYes{See Section \ref{sec:conclusion}}
  \item Did you discuss any potential negative societal impacts of your work?
    \answerYes{See Section \ref{sec:conclusion}}
  \item Have you read the ethics review guidelines and ensured that your paper conforms to them?
    \answerYes{}
\end{enumerate}

\item If you are including theoretical results...
\begin{enumerate}
  \item Did you state the full set of assumptions of all theoretical results?
    \answerYes{}
        \item Did you include complete proofs of all theoretical results?
    \answerYes{}
\end{enumerate}

\item If you ran experiments...
\begin{enumerate}
  \item Did you include the code, data, and instructions needed to reproduce the main experimental results (either in the supplemental material or as a URL)?
    \answerYes{See Section \ref{sec:experimental_details}}
  \item Did you specify all the training details (e.g., data splits, hyperparameters, how they were chosen)?
    \answerYes{See Section \ref{sec:experimental_details}}
        \item Did you report error bars (e.g., with respect to the random seed after running experiments multiple times)?
    \answerYes{See Section \ref{sec:experimental_details}}
        \item Did you include the total amount of compute and the type of resources used (e.g., type of GPUs, internal cluster, or cloud provider)?
    \answerNo{Our method is not resource hungry, so there's no need to specify GPU type and computation type.}
\end{enumerate}

\item If you are using existing assets (e.g., code, data, models) or curating/releasing new assets...
\begin{enumerate}
  \item If your work uses existing assets, did you cite the creators?
    \answerYes{}
  \item Did you mention the license of the assets?
    \answerNo{The licences are mentioned in cited papers.}
  \item Did you include any new assets either in the supplemental material or as a URL?
    \answerNo{}
  \item Did you discuss whether and how consent was obtained from people whose data you're using/curating?
    \answerNo{Those are discussed in cited papers.}
  \item Did you discuss whether the data you are using/curating contains personally identifiable information or offensive content?
    \answerNo{Those are discussed in cited papers.}
\end{enumerate}

\item If you used crowdsourcing or conducted research with human subjects...
\begin{enumerate}
  \item Did you include the full text of instructions given to participants and screenshots, if applicable?
    \answerNA{We did not use crowdsourcing nor conduct research with human subjects.}
  \item Did you describe any potential participant risks, with links to Institutional Review Board (IRB) approvals, if applicable?
    \answerNA{We did not use crowdsourcing nor conduct research with human subjects.}
  \item Did you include the estimated hourly wage paid to participants and the total amount spent on participant compensation?
    \answerNA{We did not use crowdsourcing nor conduct research with human subjects.}
\end{enumerate}

\end{enumerate}

%%%%%%%%%%%%%%%%%%%%%%%%%%%%%%%%%%%%%%%%%%%%%%%%%%%%%%%%%%%%

\clearpage

\appendix

% \section{Appendix}

% Optionally include extra information (complete proofs, additional experiments and plots) in the appendix.
% This section will often be part of the supplemental material.

\end{document}